\documentclass[letterpaper]{article} 
\usepackage{aaai2026}  
\usepackage{times}  
\usepackage{helvet}  
\usepackage{courier}  
\usepackage[hyphens]{url}  
\usepackage{graphicx} 
\usepackage{natbib}  
\usepackage{caption} 
\usepackage{algorithm}
\usepackage{algpseudocode}
\usepackage{amsmath}
\usepackage{amsfonts}
\usepackage{xcolor}
\usepackage{booktabs}
\usepackage{subfig}
\usepackage{newfloat}
\usepackage{listings}
\DeclareCaptionStyle{ruled}{labelfont=normalfont,labelsep=colon,strut=off} 
\floatstyle{ruled}
\newfloat{listing}{tb}{lst}{}
\floatname{listing}{Listing}
\nocopyright 

\title{Rethinking Random Masking in Self-Distillation on ViT}
\author{
    Jihyeon Seong\textsuperscript{\rm 1}\equalcontrib,
    Hyunkyung Han\textsuperscript{\rm 2}\equalcontrib
}
\affiliations{
    \textsuperscript{\rm 1}Korea Advanced Institute of Science and Technology (KAIST), South Korea\\
    \textsuperscript{\rm 2}Yonsei University, South Korea\\
    jihyeon.seong@kaist.ac.kr, stellahan@yonsei.ac.kr
}

\begin{document}

\maketitle

\begin{abstract}
Vision Transformers (ViTs) have demonstrated remarkable performance across a wide range of vision tasks. In particular, self-distillation frameworks such as DINO have contributed significantly to these advances. Within such frameworks, random masking is often utilized to improve training efficiency and introduce regularization. However, recent studies have raised concerns that indiscriminate random masking may inadvertently eliminate critical semantic information, motivating the development of more informed masking strategies. In this study, we explore the role of random masking in the self-distillation setting, focusing on the DINO framework. Specifically, we apply random masking exclusively to the student’s global view, while preserving the student’s local views and the teacher’s global view in their original, unmasked forms. This design leverages DINO’s multi-view augmentation scheme to retain clean supervision while inducing robustness through masked inputs. We evaluate our approach using DINO-Tiny on the mini-ImageNet dataset and show that random masking under this asymmetric setup yields more robust and fine-grained attention maps, ultimately enhancing downstream performance.
\end{abstract}


\section{Introduction}
Vision Transformers (ViTs) have demonstrated impressive performance across a variety of visual recognition tasks\cite{naseer2021intriguing}. However, their success has traditionally depended on large-scale labeled datasets, as ViTs tend to underperform when trained on limited data due to the lack of strong inductive biases. To mitigate this limitation, knowledge distillation techniques have been employed, wherein a compact ViT (student) learns from the soft predictions or internal representations of a large, pretrained model (teacher), thereby significantly improving the student’s generalization. Nonetheless, this approach inherently relies on the availability of a powerful teacher model trained on massive datasets, which can be impractical in many settings. To address this issue, self-distillation has emerged as a compelling alternative: it removes the dependence on external teacher models by training a student and teacher of identical architecture jointly\cite{shen2022self}. In many self-distillation frameworks, the teacher is updated using an exponential moving average (EMA) of the student’s weights, enabling the model to benefit from its own past knowledge in a stable and gradually evolving manner\cite{he2020momentum, grill2020bootstrap}.

In knowledge distillation, random masking has been adopted as a strategy to improve the robustness and generalization of the student model\cite{zhu2018knowledge}. Under this framework, the teacher model remains fixed and processes full input images, while the student is trained on randomly masked inputs, where a subset of image patches is removed before encoding. Despite receiving only partial observations, the student is supervised to match the teacher’s output, thereby learning to infer discriminative representations from incomplete data. However, recent studies have pointed out that random masking may inadvertently discard semantically important regions, which are crucial for effective supervision. To address this limitation, guided masking strategies have been proposed, where the masking process is informed by attention maps, saliency scores, or learned importance weights. By selectively masking less informative regions, these approaches ensure that the student observes and learns from the most relevant content, leading to more stable and effective knowledge transfer.

While masking strategies have been extensively explored in knowledge distillation frameworks—often to regularize the student by exposing it to incomplete inputs—their application in self-distillation remains relatively underexplored. In self-distillation, where both teacher and student share the same architecture and are trained jointly, most research has focused on modifying training objectives, loss balancing, or architecture design, rather than input-level transformations. Although some recent works, such as iBOT and DINOv2, have implicitly introduced masking by applying asymmetric views of the input image, the teacher and student are trained simultaneously, making it unclear how random masking affects their interaction and mutual representation learning. As a result, the impact of masking in self-distillation remains an open question, warranting further investigation into how partial inputs influence both branches during joint optimization\cite{zhou2021ibot, oquab2023dinov2}.

In this study, we investigate the effect of random masking in self-distillation, particularly within the DINO framework. Leveraging the view augmentation scheme of DINO, we apply random masking only to the student's global view, while keeping the local cropped views and the teacher's global view unmasked. This design enables us to isolate and assess the impact of masking under asymmetric input conditions. We evaluate how such view-specific random masking influences the performance of DINO. Experiments are conducted using DINO-tiny on the mini-ImageNet dataset.

\section{Related Work}
Knowledge distillation (KD) and self-distillation are two related but distinct paradigms for transferring knowledge to improve model performance. In knowledge distillation, a pretrained, fixed teacher model, typically larger or more powerful, guides the learning of a student model by providing soft targets or intermediate representations\cite{hinton2015distilling}. This setup often assumes that the teacher is trained on large-scale data and remains unchanged during student training. In contrast, self-distillation removes the dependency on an external teacher by using models of the same architecture trained jointly, where one model acts as a temporal or structural teacher\cite{yuan2020bote, zhang2019self}. Notably, methods like DINO\cite{caron2021emerging} and BYOT\cite{grill2020bootstrap} employ self-distillation where the teacher is updated as an exponential moving average (EMA) of the student, allowing knowledge to accumulate over time. While knowledge distillation emphasizes transferring prior expertise, self-distillation focuses on bootstrapping knowledge within a single model framework, offering a more data-efficient and self-contained alternative.

In the context of KD for Vision Transformers, masking strategies have been employed to improve training efficiency and student robustness. Several methods adopt random masking, where a subset of image patches is randomly removed from the student's input while the teacher processes the full image. For instance, MaskedKD introduces random masking to the teacher’s input to reduce computational cost, and iBOT applies random crops and masks to the student's view to enforce prediction consistency across asymmetric inputs. These random masking strategies act as regularizers but may obscure semantically important regions. To address this, recent works propose guided masking, where the masking decision is informed by semantic relevance or attention scores. In TokenDrop, less informative tokens—determined by attention magnitudes—are dropped before distillation, preserving critical content for learning. Similarly, Saliency-guided KD uses saliency maps to retain important visual regions while masking the rest. These guided approaches aim to selectively remove uninformative content, ensuring the student receives meaningful supervision while still benefiting from reduced redundancy and enhanced generalization.

While masking strategies have been actively studied in knowledge distillation frameworks, their role in self-distillation remains relatively underexplored. Some recent self-distillation methods, such as DINO and iBOT, implicitly incorporate random masking through view augmentation, where the student receives a randomly cropped or masked global view, while the teacher operates on a cleaner or differently augmented input. Although this introduces asymmetry that benefits representation learning, the masking is typically applied in a random and heuristic manner, without principled analysis of its impact on learning dynamics or supervision quality. Unlike knowledge distillation, where guided masking has been developed to preserve semantically important regions, self-distillation has yet to explore how masking strategies—random or guided—can be systematically optimized. This presents a promising research direction for improving the effectiveness and interpretability of self-distillation frameworks in Vision Transformers.

\begin{algorithm}[H]
\caption{Random Masking in DINO Self-Distillation (Student Global View Only)}
\label{alg:random_masking}
\begin{algorithmic}[1]
\Require Image tensor $X \in \mathbb{R}^{B \times C \times H \times W}$, patch size $p$, mask ratio $r$, grid size $(g_H, g_W)$
\Ensure Masked image tensor $\tilde{X} \in \mathbb{R}^{B \times C \times H \times W}$
\State $P \gets g_H \times g_W$ \Comment{Total number of patches}
\State $k \gets \lfloor r \cdot P \rfloor$ \Comment{Number of patches to mask}
\For{$i = 1$ to $B$}
    \State $S_i \sim \text{UniformSample}(P, k)$ \Comment{Sample $k$ patch indices uniformly}
    \State $M_i \gets \mathbf{1} \in \{0,1\}^{g_H \times g_W}$ \Comment{Initialize binary mask}
    \For{each $j \in S_i$}
        \State Set $M_i[j] \gets 0$ \Comment{Mask selected patch index}
    \EndFor
    \State $\hat{M}_i \gets \text{NearestInterp}(M_i, H, W)$ \Comment{Upsample}
    \State $\tilde{X}_i \gets X_i \cdot \hat{M}_i$ \Comment{Apply pixel-level mask}
\EndFor
\State \Return $\tilde{X} = \{\tilde{X}_i\}_{i=1}^{B}$
\end{algorithmic}
\end{algorithm}

\begin{figure}[h]
    \centering
    \includegraphics[width=0.5\textwidth]{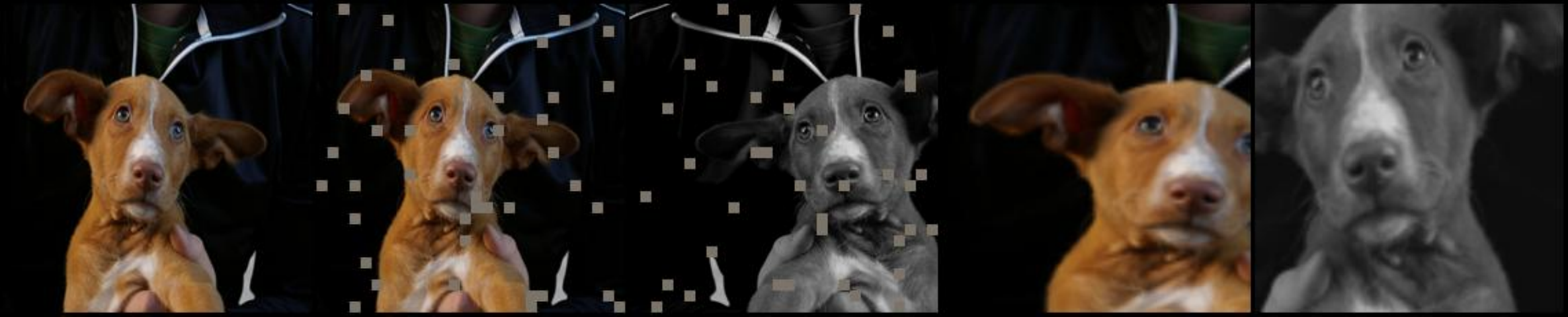}
    \caption{Random Masking Example: 1-original image, 2-first global view masked image, 3-second global view masked image, 4-first local view clean image, 5-second local view clean image}
    \label{fig:masking}
\end{figure}

\begin{table}[ht]
    \centering
    \begin{tabular}{c|cc}
    \hline
        Type & k-NN & Lin. Prob \\
        \hline
        DINO-tiny & 53.02 & 59.53 \\
        DINO-tiny w/ R.M & 53.01 & 60.29\\
    \hline
    \end{tabular}
    \caption{Top-1 Accuracy between Vanilla DINO and Randomly Masked DINO on mini-ImageNet.}
    \label{tab:performance}
\end{table}

\section{Random Masking Analysis}
We specifically analyze the role of random masking within the augmented view in self-distillation. To isolate its effect, we apply masking only to the student’s global view, while retaining clean versions for both the student’s local views and the teacher’s global view. As shown in  Algorithm~\ref{alg:random_masking}, random masking is applied during training to enhance robustness and mitigate overfitting. Each image is divided into a grid of non-overlapping patches, determined by the patch size and image resolution. A fixed masking ratio $r$ is used to calculate the number of patches to be removed. For every image in the batch, a subset of patch indices is sampled uniformly at random, without relying on semantic cues. A binary patch-level mask is constructed by setting the selected indices to zero (masked) and the others to one (visible). This mask is then upsampled to full resolution using nearest-neighbor interpolation to obtain a pixel-level mask with sharp boundaries. The masked image is computed by element-wise multiplication of the binary mask and the original image tensor. This masked global view is provided solely to the student model, while the teacher operates on a clean global view, enforcing cross-view consistency and serving as a regularization mechanism in the self-distillation process.

The key difference between the random masking used in our DINO setup and the patch dropping in DINOv2 lies in where and how the masking is applied. In our approach, random masking is performed at the input image level, where selected patches are zeroed out before being passed to the student model, creating an explicit asymmetry between the masked student view and the clean teacher view. This acts as a form of denoising regularization, encouraging the student to align with the teacher despite partial information. In contrast, DINOv2 applies patch dropping internally by discarding a subset of tokens after the patch embedding stage, without modifying the raw input or reconstructing the missing tokens\cite{oquab2023dinov2}. This technique primarily serves to reduce computational cost and acts as an implicit regularizer, rather than enforcing a denoising task. Thus, while both methods reduce visible information, they differ in mechanism, timing, and training objective.

\begin{table}[t]
\small  
\centering
\begin{tabular}{@{}ll@{}}
\toprule
\textbf{Component} & \textbf{Setting} \\
\midrule
Architecture & ViT-Tiny, patch size 16 \\
Epochs & 100 \\
Batch size & 50 per GPU \\
Optimizer & AdamW \\
Learning rate & 0.0005 (scaled) \\
Weight decay & 0.04 $\rightarrow$ 0.4 (cosine) \\
Grad clip & 3.0 \\
FP16 & Enabled \\
\midrule
Teacher EMA & 0.996 $\rightarrow$ 1.0 \\
DINO loss dim & 65,536 \\
Temp (S/T) & 0.1 / 0.04 \\
Center momentum & 0.9 \\
\midrule
Global crops & 2 (scale: 0.4–1.0) \\
Local crops & 8 (scale: 0.05–0.4) \\
\bottomrule
\end{tabular}
\caption{DINO hyperparameters (ViT-Tiny)}
\label{tab:dino_hyperparams}
\end{table}

\subsection{Experimental Setup}
We train DINO-Tiny on the mini-ImageNet dataset using two NVIDIA RTX 3090 GPUs with PyTorch and CUDA 12.2. The detailed hyperparameter settings are summarized in Table~\ref{tab:dino_hyperparams}. For the random masking experiments, we apply a masking ratio of 0.1, meaning 10\% of the image patches in the student’s global view are randomly masked during training.

\begin{figure}
    \centering
    \subfloat[Original Img]{\includegraphics[width=0.13\textwidth]{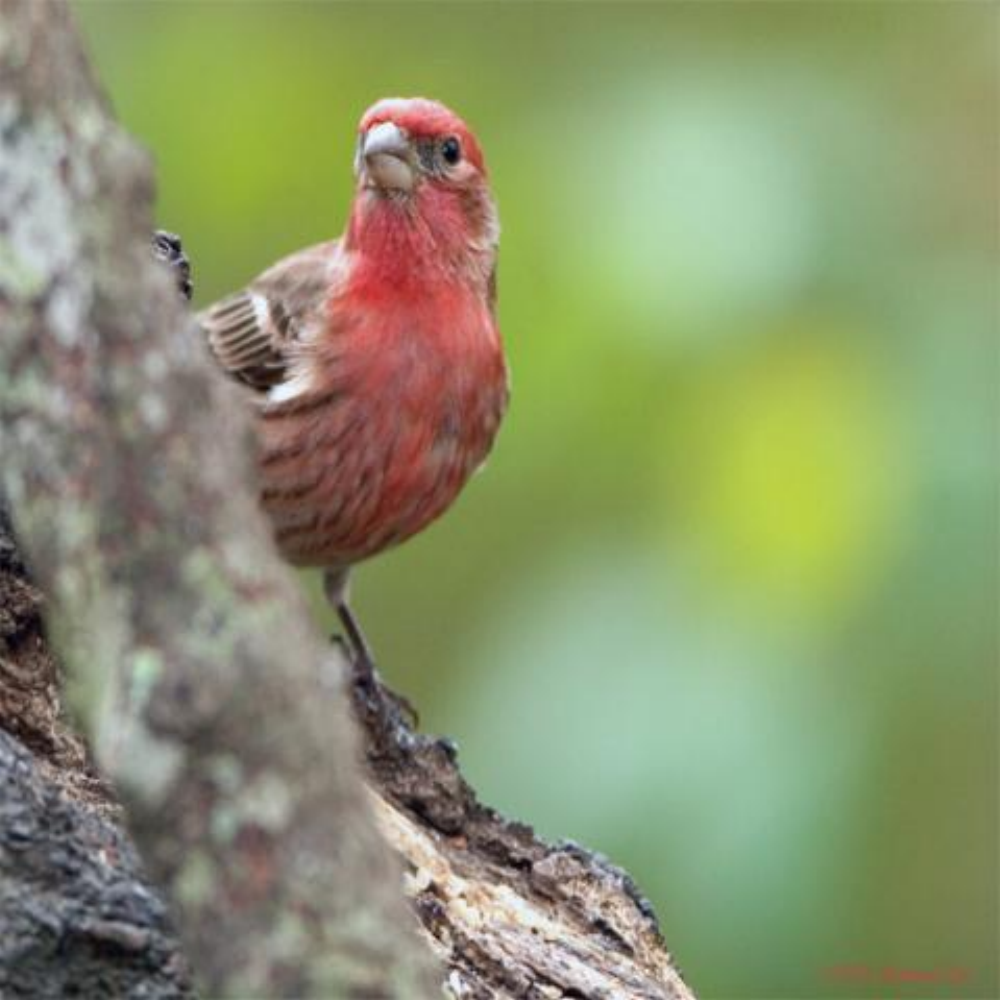}}
    \subfloat[DINO]{\includegraphics[width=0.13\textwidth]{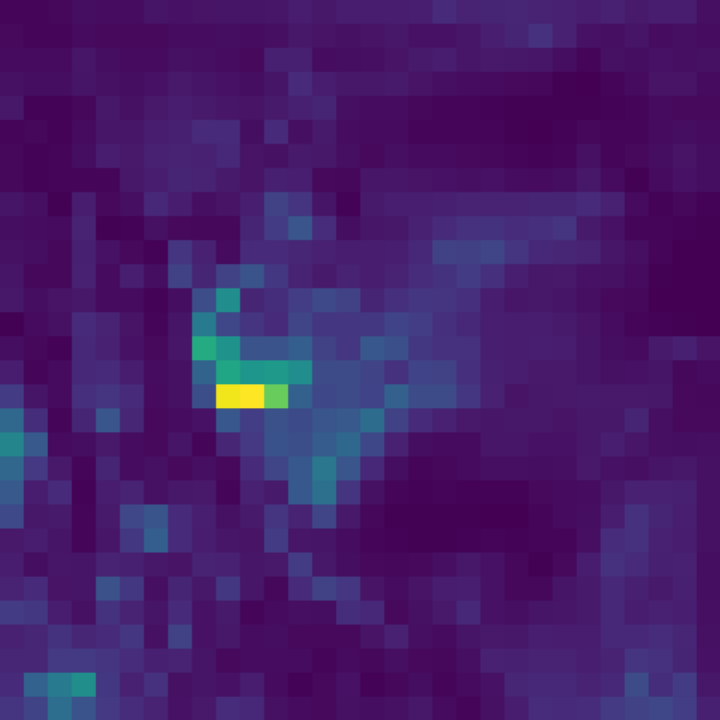}}
    \subfloat[DINO w/ R.M]{\includegraphics[width=0.13\textwidth]{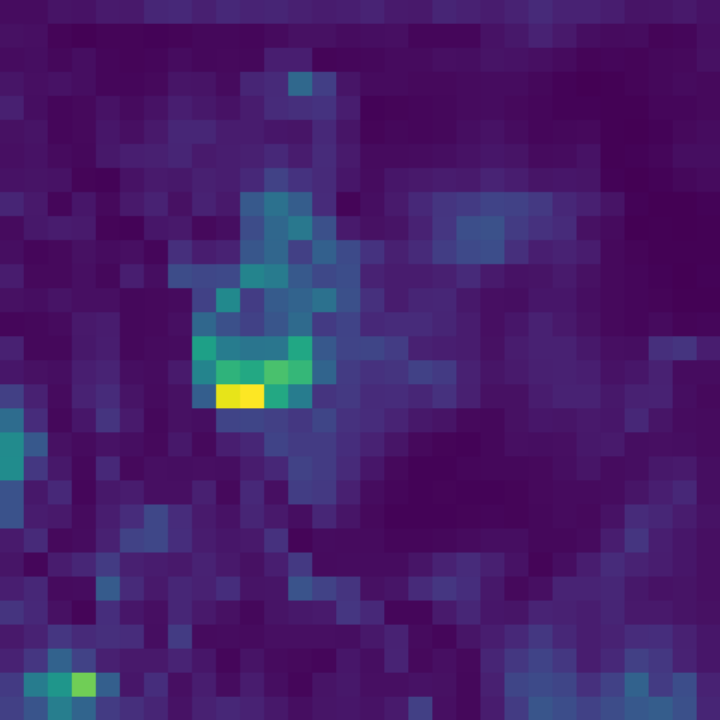}}
    \caption{Attention Visualization}
    \label{fig:attn}
\end{figure}
\subsection{Evaluation Metrics}
We evaluate the learned representations on mini-ImageNet using two standard protocols: $k$-Nearest Neighbors ($k$-NN) and linear probing, both reporting top-1 accuracy. The $k$-NN classifier allows us to assess how fine-grained and instance-level the ViT's learned features are, without additional parameter tuning. In contrast, linear probing measures the linear separability of the learned representations by training a single fully connected layer on top of the frozen backbone, offering a stronger signal of how well the model captures class-level discriminative structure.

\subsection{Results}
As shown in Table \ref{tab:performance}, while the k-NN accuracy exhibits only a slight degradation, linear probing performance improves noticeably. This suggests that the learned representations become more linearly separable, indicating enhanced feature quality despite the randomness introduced during training. Furthermore, as illustrated in Figure \ref{fig:attn}, the attention maps generated by our random masking display more focus on the bird object compared to the vanilla-DINO. This observation highlights the potential of random masking, when applied asymmetrically within the self-distillation pipeline, to enhance the robustness and localization capabilities of Vision Transformers.

\section{Conclusion}
In this study, we investigated the impact of random masking applied exclusively to the student’s global view in the self-distillation setting of DINO. By preserving clean views for the teacher and the students’ local crops, we isolated the effect of view-specific partial observation on representation learning. Our experiments on mini-ImageNet using DINO-Tiny demonstrate that even lightweight random masking introduces meaningful regularization, influencing both fine-grained retrieval (evaluated via $k$-NN) and class-level separability (via linear probing). While random masking offers performance gains under certain conditions, our results also highlight the need for a deeper understanding of masking dynamics in self-distillation, suggesting that more structured or guided masking strategies could further enhance representation quality. This work opens the door to future research on adaptive masking techniques tailored specifically for self-distillation frameworks.

\section{Acknowledgement}
This work was partially supported by an IITP grant funded by the Korean Government(MSIT) (No. RS-2020-II201361, Artificial Intelligence Graduate School Program (Yonsei University))

\bibliography{aaai2026}

\begin{thebibliography}{11}
\providecommand{\natexlab}[1]{#1}

\bibitem[{Caron et~al.(2021)Caron, Touvron, Misra, J{\'e}gou, Mairal, Bojanowski, and Joulin}]{caron2021emerging}
Caron, M.; Touvron, H.; Misra, I.; J{\'e}gou, H.; Mairal, J.; Bojanowski, P.; and Joulin, A. 2021.
\newblock Emerging Properties in Self-Supervised Vision Transformers.
\newblock \emph{Proceedings of the IEEE/CVF International Conference on Computer Vision}, 9650--9660.

\bibitem[{Grill et~al.(2020)Grill, Strub, Moraldo, Masurel, Schmid, Altche, Tallec, Richemond, Savalle, Doersch et~al.}]{grill2020bootstrap}
Grill, J.-B.; Strub, F.; Moraldo, F.; Masurel, P.; Schmid, A.; Altche, G.; Tallec, C.; Richemond, A.; Savalle, E.; Doersch, C.; et~al. 2020.
\newblock Bootstrap Your Own Latent--A New Approach to Self-Supervised Learning.
\newblock \emph{Advances in Neural Information Processing Systems}, 33: 17849--17861.

\bibitem[{He et~al.(2020)He, Fan, Xie, Girshick, and Doll{\'a}r}]{he2020momentum}
He, K.; Fan, H.; Xie, Y.; Girshick, R.; and Doll{\'a}r, P. 2020.
\newblock Momentum Contrast for Unsupervised Visual Representation Learning.
\newblock \emph{Proceedings of the IEEE/CVF Conference on Computer Vision and Pattern Recognition}, 9729--9738.

\bibitem[{Hinton, Vinyals, and Dean(2015)}]{hinton2015distilling}
Hinton, G.; Vinyals, O.; and Dean, J. 2015.
\newblock Distilling the knowledge in a neural network.
\newblock \emph{arXiv preprint arXiv:1503.02531}.

\bibitem[{Naseer et~al.(2021)Naseer, Ranasinghe, Khan, Hayat, Shahbaz~Khan, and Yang}]{naseer2021intriguing}
Naseer, M.~M.; Ranasinghe, K.; Khan, S.~H.; Hayat, M.; Shahbaz~Khan, F.; and Yang, M.-H. 2021.
\newblock Intriguing properties of vision transformers.
\newblock \emph{Advances in Neural Information Processing Systems}, 34: 23296--23308.

\bibitem[{Oquab et~al.(2023)Oquab, Darcet, Moutakanni, Vo, Szafraniec, Khalidov, Fernandez, Haziza, Massa, El-Nouby et~al.}]{oquab2023dinov2}
Oquab, M.; Darcet, T.; Moutakanni, T.; Vo, H.; Szafraniec, M.; Khalidov, V.; Fernandez, P.; Haziza, D.; Massa, F.; El-Nouby, A.; et~al. 2023.
\newblock Dinov2: Learning robust visual features without supervision.
\newblock \emph{arXiv preprint arXiv:2304.07193}.

\bibitem[{Shen et~al.(2022)Shen, Xu, Yang, Li, and Guo}]{shen2022self}
Shen, Y.; Xu, L.; Yang, Y.; Li, Y.; and Guo, Y. 2022.
\newblock Self-distillation from the last mini-batch for consistency regularization.
\newblock In \emph{Proceedings of the IEEE/CVF conference on computer vision and pattern recognition}, 11943--11952.

\bibitem[{Yuan et~al.(2020)Yuan, Chen, Wang, Li, Xue, Li, and Li}]{yuan2020bote}
Yuan, L.; Chen, B.; Wang, M.; Li, W.; Xue, G.; Li, F.; and Li, W. 2020.
\newblock Knowledge Distillation by On-the-Fly Native Ensemble.
\newblock \emph{Advances in Neural Information Processing Systems}, 33: 15589--15600.

\bibitem[{Zhang and Wu(2019)}]{zhang2019self}
Zhang, H.; and Wu, J. 2019.
\newblock Self-Distillation from the Last Mini-Batch.
\newblock \emph{arXiv preprint arXiv:1907.03964}.

\bibitem[{Zhou et~al.(2021)Zhou, Wei, Wang, Shen, Xie, Yuille, and Kong}]{zhou2021ibot}
Zhou, J.; Wei, C.; Wang, H.; Shen, W.; Xie, C.; Yuille, A.; and Kong, T. 2021.
\newblock ibot: Image bert pre-training with online tokenizer.
\newblock \emph{arXiv preprint arXiv:2111.07832}.

\bibitem[{Zhu, Gong et~al.(2018)}]{zhu2018knowledge}
Zhu, X.; Gong, S.; et~al. 2018.
\newblock Knowledge distillation by on-the-fly native ensemble.
\newblock \emph{Advances in neural information processing systems}, 31.

\end{thebibliography}

\end{document}